\documentclass[conference,10pt,a4paper]{IEEEtran}  
\IEEEoverridecommandlockouts
\usepackage{cite}
\usepackage{amsmath,amssymb,amsfonts}
\usepackage{algorithmic}
\usepackage{graphicx}
\usepackage{rotating}
\usepackage[leftcaption]{sidecap}
\usepackage{textcomp}
\usepackage[table,dvipsnames]{xcolor}
\usepackage{etoolbox}

\graphicspath{{figures/}}

\usepackage{listings}
\definecolor{codegray}{rgb}{0.5,0.5,0.5}

\definecolor{maroon}{rgb}{0.5,0,0}
\definecolor{darkgreen}{rgb}{0,0.5,0}

\lstdefinelanguage{XML}
{
  basicstyle=\ttfamily\tiny, 
  morestring=[s]{"}{"},
  morecomment=[s]{?}{?},
  morecomment=[s]{!--}{--},
  commentstyle=\color{darkgreen},
  moredelim=[s][\color{black}]{>}{<},
  moredelim=[s][\color{red}]{\ }{=},
  stringstyle=\color{blue},
  identifierstyle=\color{maroon}
}

\ifCLASSOPTIONcompsoc
\usepackage[caption=false,font=normalsize,labelfont=sf,textfont=sf]{subfig}
\else
\usepackage[caption=false,font=footnotesize]{subfig}
\fi

\def\BibTeX{{\rm B\kern-.05em{\sc i\kern-.025em b}\kern-.08em
    T\kern-.1667em\lower.7ex\hbox{E}\kern-.125emX}}
    
\usepackage{algorithmic}

\usepackage{tikz}

\usepackage{verbatim}

\begin{document}

\title{An Integrated Toolbox for Creating Neuromorphic Edge Applications}

\author{\IEEEauthorblockN{
Lars Niedermeier\IEEEauthorrefmark{1}
and Jeffrey L. Krichmar\IEEEauthorrefmark{2}}
\IEEEauthorblockA{\IEEEauthorrefmark{1}\textit{Niedermeier Consulting, Zurich, ZH, Switzerland}}
\IEEEauthorblockA{\IEEEauthorrefmark{2}\textit{Department of Cognitive Sciences, Department of Computer Science}, \textit{University of California, Irvine, CA, USA}}
\IEEEauthorblockA{Correspondence Email: lars@niedermeier-consulting.ch}  
}

\maketitle

\begin{abstract}
Spiking Neural Networks (SNNs) and neuromorphic models are more efficient and have more biological realism than the activation functions typically used in deep neural networks, transformer models and generative AI. SNNs have local learning rules, are able to learn on small data sets, and can adapt through neuromodulation. Although research has shown their advantages, there are still few compelling practical applications, especially at the edge where sensors and actuators need to be processed in a timely fashion.
One reason for this might be that SNNs are much more challenging to 
understand, build, and operate due to their intrinsic properties.
For instance, the mathematical foundation involves differential equations 
rather than basic activation functions. 
To address these challenges, we have developed CARLsim++. 
It is an integrated toolbox that enables fast and easy creation of neuromorphic applications. 
It encapsulates the mathematical intrinsics and low-level C++ programming 
by providing a graphical user interface for users who do not have a background in software engineering but still want to create neuromorphic models. Developers can easily configure inputs and outputs to devices and robots. These can be accurately simulated before deploying on physical devices. CARLsim++ can lead to rapid development of neuromorphic applications for simulation or edge processing. 
\end{abstract}

\begin{IEEEkeywords}
Edge Computing,
Neuromorphic Applications,
Spiking Neural Networks 
\end{IEEEkeywords}



\section{Introduction}

Simulation systems that support Spiking Neural Networks (SNNs) are more relevant than ever for research and application development. SNN models have been used to understand the brain, accelerate machine learning algorithms with highly parallel or event-driven systems, and support neuromorphic computation. The CARLsim SNN framework was one of the first Open Source simulation systems that utilized CUDA GPUs to address the tremendous parallel processing demands of natural brains \cite{nageswaran2009}. It has evolved over a decade in numerous scientific research projects requiring efficient biologically plausible modeling at scale. Currently in its sixth major release, CARLsim 6 supports the latest versions of operating systems, development tool chains, multi-core computers, and GPUs \cite{Niedermeier2022}. 

\begin{figure*}[htb]  %
\centerline{\includegraphics[width=0.875\textwidth]{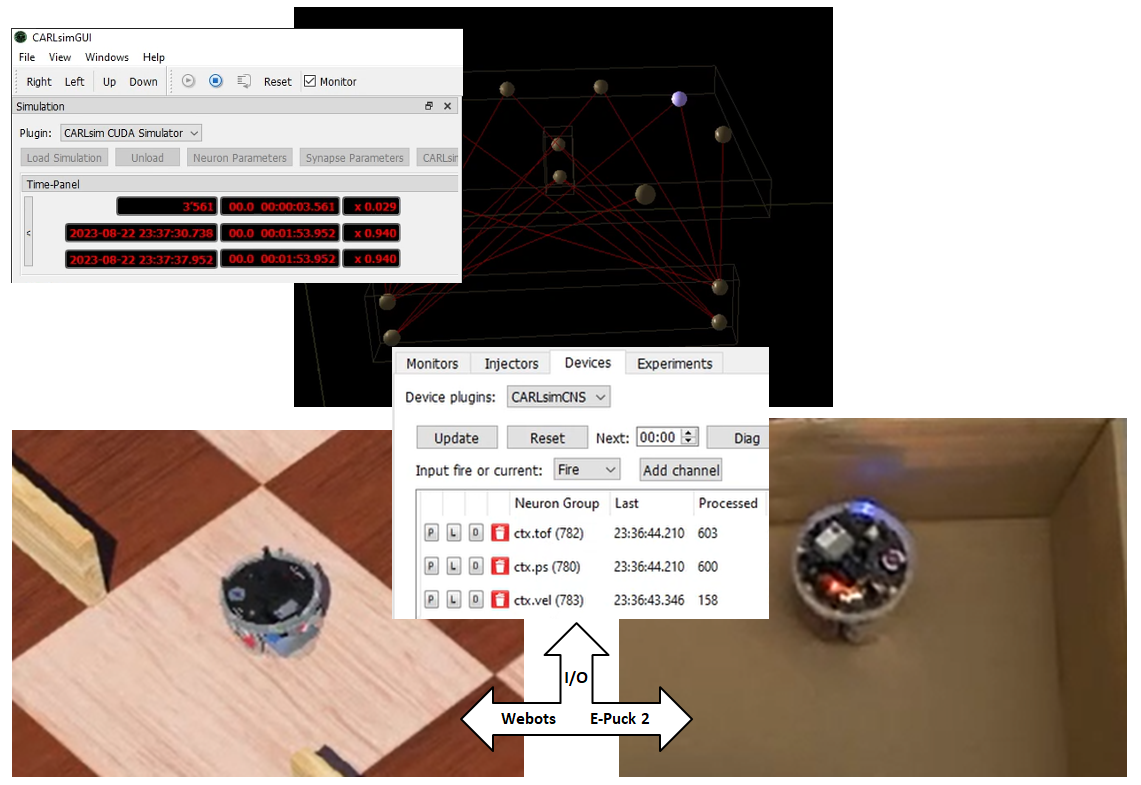}}    
\caption{CARLsim++ GUI and I/O Interface. Top: Screenshot showing the neural activity of the SNN model in the GUI. Left: Webots simulation of E-Puck robot controlled by CARLsim I/O interface. Center: I/O interface shows listing channels for the E-Puck's sensors and actuators. Right: Porting model to physical E-Puck robot as it detects a wall by its front-right proximity sensor, and performs obstacle avoidance.}
\label{fig:io}
\end{figure*}

The goal of the present work is to extend CARLsim to allow for rapid deployment of practical applications, especially those that interface with sensors and actuators in real-time. We believe there is a need for an easy to use SNN platform that supports application development of large-scale, detailed models. A roadblock for the community has been difficult interfaces and installation requirements. Furthermore, users want to incorporate new neuron models, synapse models, and learning rules in their simulations supported by an application framework. We demonstrate these capabilities with a neurorobotic application using the E-Puck robot (Fig. \ref{fig:io}).
%


%
%

%


\color{black}

CARLsim++ provides the required tools in an integrated framework,   
so that neuroscientists, roboticists, and embedded system developers can focus on the creation of neuromorphic systems on real world data with practical applications.  
Our work comprises the following contributions:
\begin{enumerate}
    \item \textbf{\textit{Tool to build neuromorphic applications.}} We demonstrate new input/output (I/O) and GUI components for CARLsim in a neurorobotics application.
    \item \textbf{\textit{I/O interfaces for integration.}} Developers want to interface their SNNs with sensors, actuators, and robots. To address this need, we developed I/O interfaces by implementing a highly integrated CARLsim layer built on the robotics middleware YARP \cite{metta2006, fitzpatrick2008, paikan2014, paikan2015, domenichelli2018}. 
    \item \textbf{\textit{Graphical user interface (GUI) for SNN development.}} Researchers often want to build machine learning models without programming.  Moreover, some programmers prefer a GUI to command-line tools. To address these needs, we have developed a GUI for model creation, execution, visualization, and analysis.
    \item \textbf{\textit{Further enhancements.}} In addition to the I/O interface and GUI, we extended CARLsim's open source library with: (i) Spike coding and decoding for the I/O interface module, (ii) Integration of the GUI module as a plug-in, and (iii) Other improvements based on user feedback.
\end{enumerate}

CARLsim++ is publicly available as open source under the MIT license at: https://github.com/UCI-CARL/CARLsim6

\section{Related Work}

Recently, members of the CARLsim team participated in a short version of the NSF I-Corps program at UCI Beall Applied Innovation. 
As a part of this program, we interviewed neuroscientists, AI/ML researchers, neuromorphic hardware designers, and roboticists. Many developers want seamless I/O interfaces to their models. Roboticists have to control actuators and receive input from multiple sensors in real-time \cite{bekey2017autonomous}. Many neuromorphic engineers develop special sensors like the DVS and DAVIS neuromorphic vision sensors that need to interface with their SNNs \cite{brandli2014,delbruck2010}. To meet this need, we have created a CARLsim library, which will be described below, to facilitate I/O with embedded systems. As a result, the present work introduces a GUI for creating and simulating SNNs. Moreover, the GUI can configure inputs and outputs to sensors and actuators for simulation and deployment.

Imperial College developed SpikeStream, a GUI focused modeling environment supporting the NeMo SNN simulator \cite{gamez2007}. 
The simulation and analysis utilized a relational database. 
In contrast, CARLsim GUI directly supports the CARLsim Offline Analysis Tool (OAT) for a data driven approach that is typical in neuroscience. 
OAT can be configured by (versioned) XMLs and be started and stopped interactively. 

Obstacle avoidance has been developed on several navigating neuromorphic platforms \cite{beyeler2015, galluppi2014, schoepe2019}. These applications follow biological principles inspired by insect and mammalian vision based navigation. Neuromorphic path planning and waypoint navigation have demonstrated orders of magnitude power efficiency than conventional algorithms \cite{koul2019, Hwu2018, koziol2013, fischl2017}. However, in all these demonstrations the models were manually tuned, difficult to visualize during development, and hard to replicate without access to a specific setup. CARLsim++ can streamline this development process and assist in standardizing neuromorphic models.  

AI/ML developers often use Python in large integrated development environments (IDEs) and JupyterLab \cite{Jupyter2024} for rapid prototyping or training. 
For deploying at the edge this kind of development requires specific solutions to deploy, monitor, operate, and improve the application.
In contrast, CARLsim++ integrates robots efficiently based on C++ and without a paradigm shift on the Edge.

Other SNN simulations frameworks such as Brian2 \cite{Brian2}, GeNN \cite{GeNN},  Nengo \cite{Nengo}, Next \cite{NEST}, and NEURON \cite{NeuronGPU}, 
have focused their development environments on computational neuroscience research.
In contrast, CARLsim++ aims to transfer proven research results to practical applications by providing components and the required methods out of the box.


\section{Methods}
\label{sec:methods} 

To introduce the new features of CARLsim++, we develop a neurorobotic \cite{hwu2022} application using the GUI and I/O to create a SNN that controls an autonomous, behaving robot in the real-world.

\subsection{Robotics middleware YARP}
\label{subsec:methods_yarp} 

CARLsim I/O leverages YARP to encapsulate robot controllers and other active nodes. YARP was developed for the iCub robot at the Istituto Italiano di Tecnologia (IIT) in a collaboration between the US Defense Advanced Research Projects Agency (DARPA) and the European Union with the goal of reducing infrastructure-level programming and increasing time spent on research-level programming \cite{metta2006}. IIT continued the work by making the robotics projects stable and long-lasting, without compromising the ability to constantly change sensors, actuators, processors, 
and networks \cite{fitzpatrick2008}. Over the years, more advanced functionality was added to YARP, such as data flow monitoring \cite{paikan2014}, communication channel prioritization \cite{paikan2015}, and recently, an event processing framework by the iCub facility of IIT \cite{glover2018}. YARP also provides the underlying communication protocols and compatibility to the ROS robotic operating system that has become a standard for roboticists \cite{quigley2009}.



\subsection{Build system YCM}
\label{subsec:YCM}

One key lesson learned on the iCub/YARP project was that combining the deliverables of loosely coupled research groups together 
is a critical success factor but also an unpopular task \cite{domenichelli2018}. 
Consequently,  IIT developed YCM that enable the configuration of multiple open source repositories
and ensured well defined and reproducible automatic builds, called \emph{Superbuilds}. 
IIT has spent a significant amount of additional effort to address sustainability by upstreaming their changes back to the community's CMake repository. CARLsim adopts YCM as reference implementation and provides templates to the growing user community to help them 
not being slowed down by release management issues.

\begin{figure*}[ht]
\centering
\subfloat[E-Puck facing wall]{\includegraphics[width=0.30\textwidth]{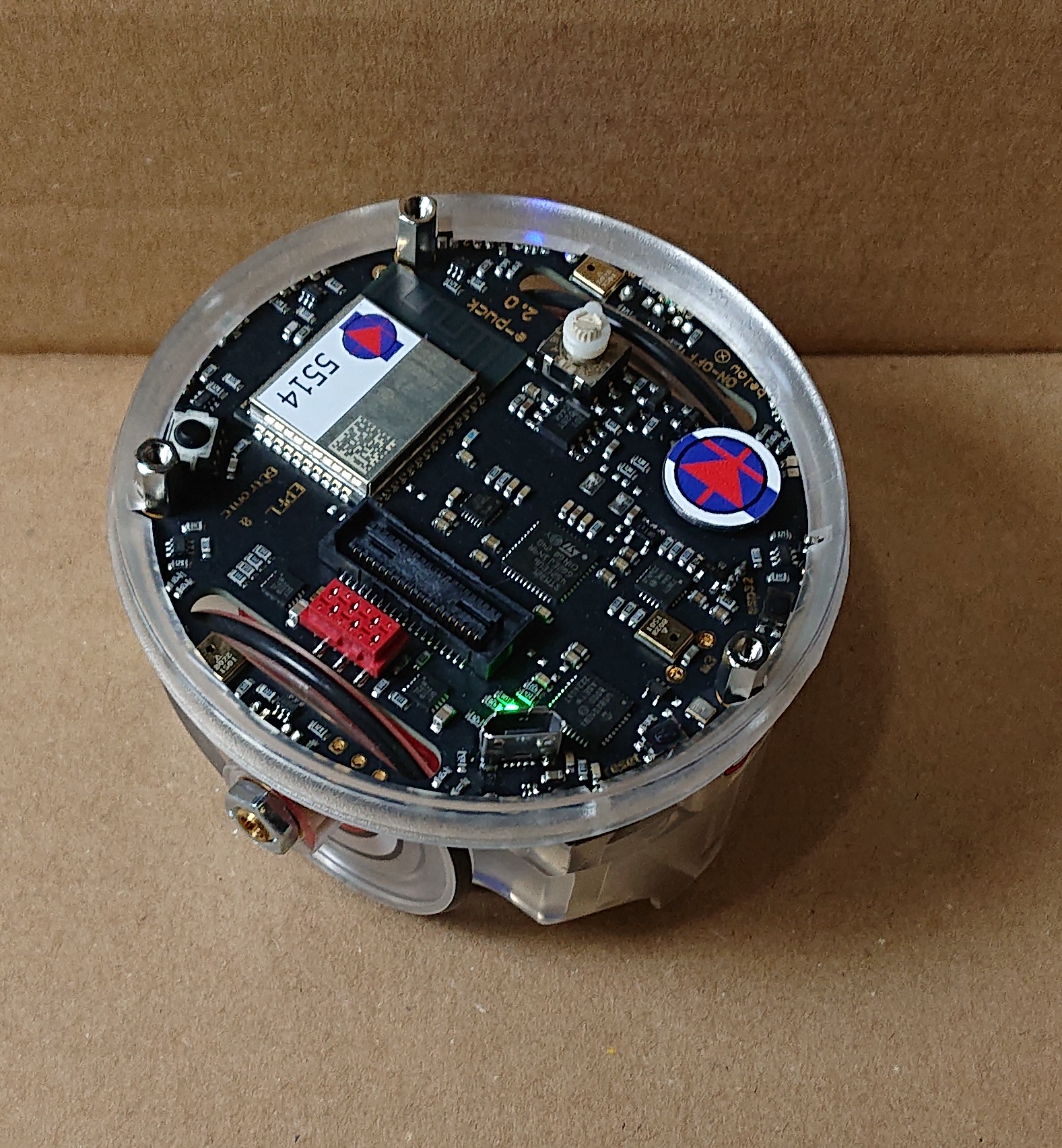}\label{subfig:EPuckWall}} 
\hfil
\subfloat[E-Puck Monitor]{\includegraphics[width=0.25\textwidth]{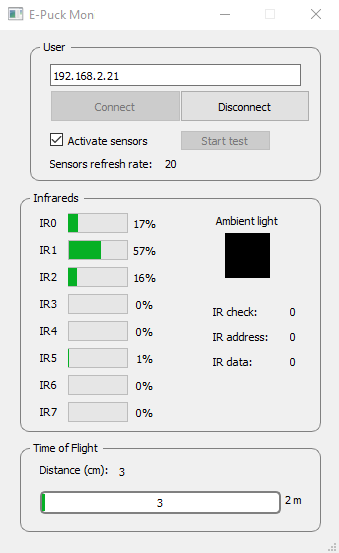}\label{subfig:EPuckMonitor}}
\hfil
\subfloat[E-Puck sensors]{\includegraphics[width=0.25\textwidth]{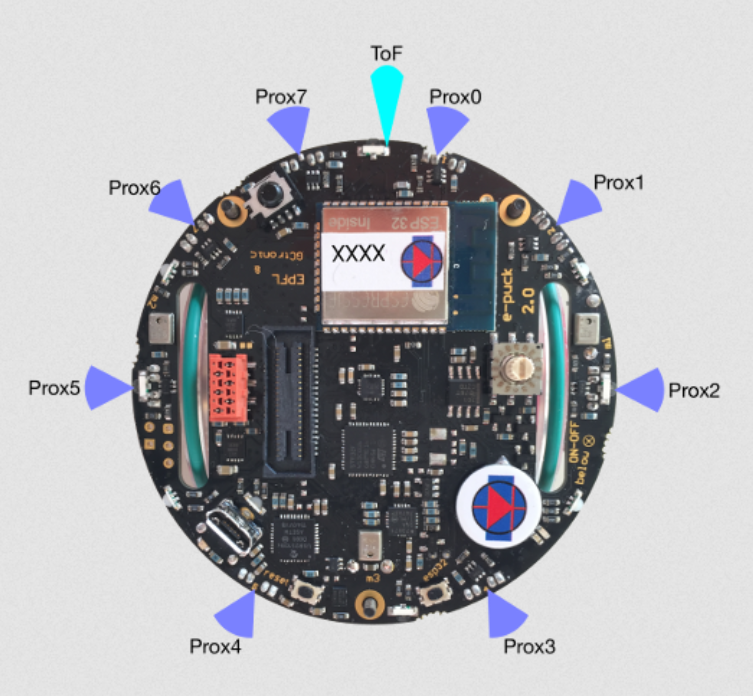}\label{subfig:EPuckSensors}}
\caption{
The E-Puck robot. 
(a) E-Puck facing a wall at the front right.
(b) E-Puck monitor application showing the corresponding sensor values. Note the high value of proximity sensor IR1. 
(c) Location of the proximity sensors. Reproduced from the E-Puck online documentation.  
}
\label{fig:EPuckWall}
\end{figure*}

\subsection{E-Puck Robot}

To demonstrate CARLsim++'s features, we present a neuromorphic robotics application that exercises the GUI and I/O interface with the E-Puck robot \cite{epuck2009}. The E-Puck is a small robot with proximity sensors and wheeled locomotion. Fig. \ref{subfig:EPuckMonitor} shows the monitor application for E-Puck that displays raw values of the sensors and access to actuators. We have used this tool for the development of the Yarp Device of the physical E-Puck. The connection is established over WiFi by providing an IP address. 

We developed a minimalistic SNN for the E-Puck robot to explore an area having boundaries and obstacles without colliding into walls or getting stuck. 
The SNN has two motor neurons for each wheel actuator. 
Their spike streams are rate coded and determine if the wheel motor rotates forward or backward. 
If the front distance sensor is facing an obstacle, the forward movement is inhibited.  
Each of the eight proximity sensors was mapped to a corresponding sensory neuron so that the robot turns away from an obstacle in its path or moves along a wall.
The minimalistic design combines the concept of Braitenberg vehicles \cite{Braitenberg1984} 
with a biological two-layered nervous system as described in Swanson 2012, Brain Architecture, Fig. 3.6 \cite{swanson2012}.
Consequently, synaptic plasticity or sophisticated optimizations were omitted. 
%
  
%




\subsection{Mathematical models}     


%
CARLsim supports both the LIF and Izhikevich models, which are implemented efficiently in CARLsim (see \cite{niedermeier2023}). Users desiring more biological realism can use the Izhikevich model without losing computational performance. In the present work, we used the Izhikevich neuron model. 
The 4-parameter model, which is dimensionless and well suited for parameter tuning, is described by the following equations \cite{Izhikevich2004}.
 
\begin{eqnarray} \label{eq:Izhi4}
\dot{v} &{}={}& 0.04v^2 + 5v + 140 - u + I    \\  
\dot{u} &{}={}& a(bv - u)  \label{eq:Izhi4u}    
\end{eqnarray}
\begin{equation} \label{eq:Izhi4v30}
\text{if}\ v > 30 \begin{cases}
v = c  \\
u = u + d
\end{cases}
\end{equation}
The dot notation indicates the differential operator $\tfrac{d}{dt}$.  
%
 
%


\color{black}

\subsection{C++ libraries}

CARLsim++ utilizes C++ libraries that encapsulate standard functionality and thus reduce the complexity of the AI application. They are well-recognized by the community, have matured over many years, and provide stable, well-tested functionality.
The CARLsim++ AI application framework exploits the following libraries.   
\emph{Boost} \cite{Boost}
is the de facto stardard for cross-platform C++ development. 
It is the spearhead of the Standard Template Library (STL), which comes with every compiler, 
and is a significant driver of C++ standardization. 
\emph{Qt} \cite{Qt6}
is designed to develop GUIs on multiple devices and platforms. 
It is also used for the KDE Linux desktop. 
\emph{SWIG} \cite{SWIG}
exposes C++ APIs to other languages such as Python or JAVA. 
\emph{MySQL}, 
\emph{Python3}, and
\emph{Matlab}/\emph{Octave}
provide the API to their namesake application as C++ libraries. 
Because the libraries are open source, seamless debugging across repository boundaries becomes possible; with stepping into the application at any arbitrary level. 
On the other hand, a consistent version level becomes a prerequisite. 
CARLsim++ addresses this challenge with the build system YCM (see \ref{subsec:YCM}). 
%

\subsection{Plug-in architecture}
\label{subsec:methods_plugins}



The plug-in architecture is a well-known design pattern to modularize complex applications without any degradation of performance. 
Plug-ins enable third-party developers to extend an application, 
support easy addition of new features, reduce the size of an application by not loading unused features, and separate source code from an application because of incompatible software licenses \cite{wikipedia_plugin}. 
At the operating system level, a plug-in is a shared library indicated by its specific extension (\emph{.dll} on Windows, \emph{.so} on Linux). 
%
%
The service defines an application programming interface (API) as contract that its plug-in has to serve. 
The plug-in manager is responsible for configuring, loading, and unloading the plug-in. 
Since a shared library is loaded into the same address space as the host application, 
the methods of the service interface are called directly by assigned function pointers. 
CARLsim++ adopts and extends the plug-in framework introduced by SpikeStream, developed at Imperial College London \cite{gamez2007}.

%
%
%
%


\section{Results}
\label{sec:results}

We present a neurorobotic demonstration that showcases CARLsim++ capabilities for developing neuromorphic applications.

\subsection{CARLsim GUI - Graphical User Interface}
\label{subsec:results_gui}

\begin{figure*}[htb]
\centering
\subfloat[Plug-in to add neuron groups]{\includegraphics[width=0.325\textwidth]{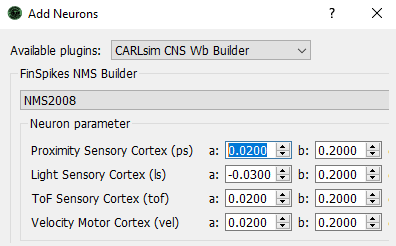}\label{subfig:addneurongroups}} 
\hfill
\subfloat[Resulting neuron groups]{\includegraphics[width=0.625\textwidth]{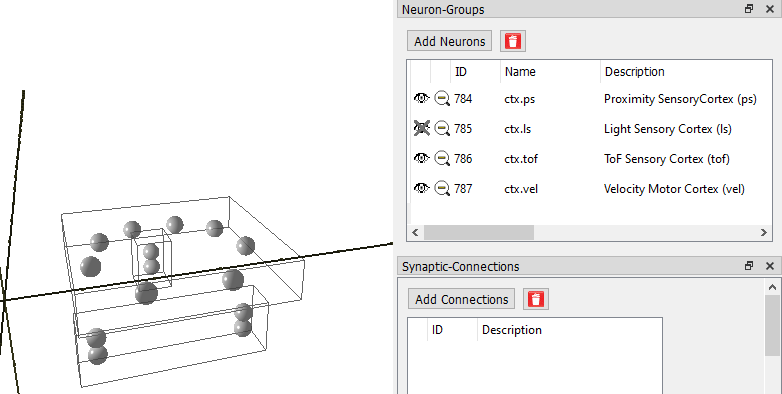}\label{subfig:neurongroups}}
\caption{
CARLsim's GUI provides plug-ins in a drop-down menu list for adding neuron groups with sensory and motor neurons.
(a) Dialog of the plug-in to add neuron groups mapped to E-Puck's proximity sensors and motor actuators.
(b) 3D Visualization and selection of sensory and motor neuron groups. 
}
\label{fig:neurons}
\end{figure*}

\begin{figure*}[htb]
\subfloat[Resulting connections]{\includegraphics[width=0.65\textwidth]{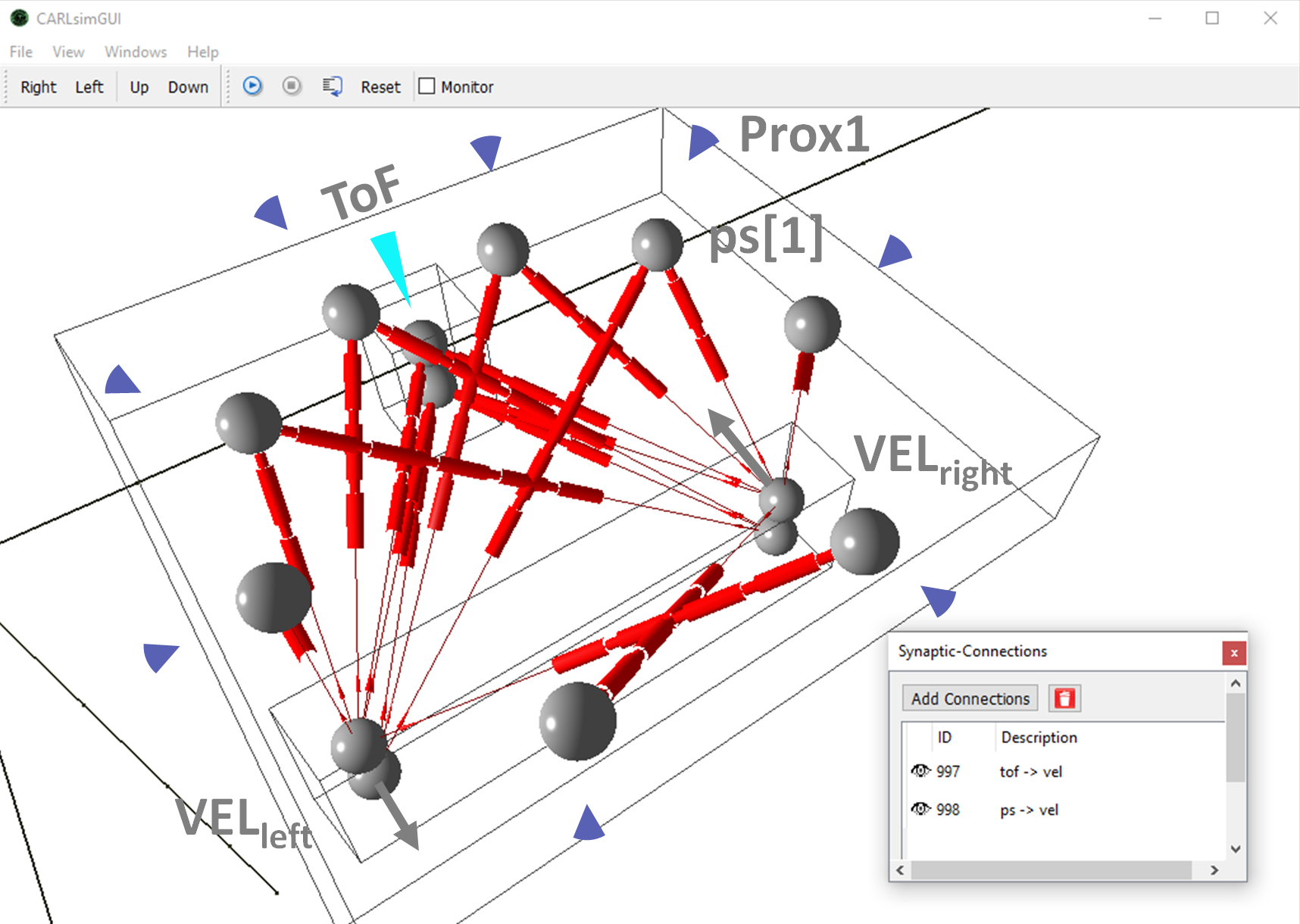}\label{subfig:cortex}}
\hfil
\subfloat[Schematic]{\includegraphics[width=0.3\textwidth]{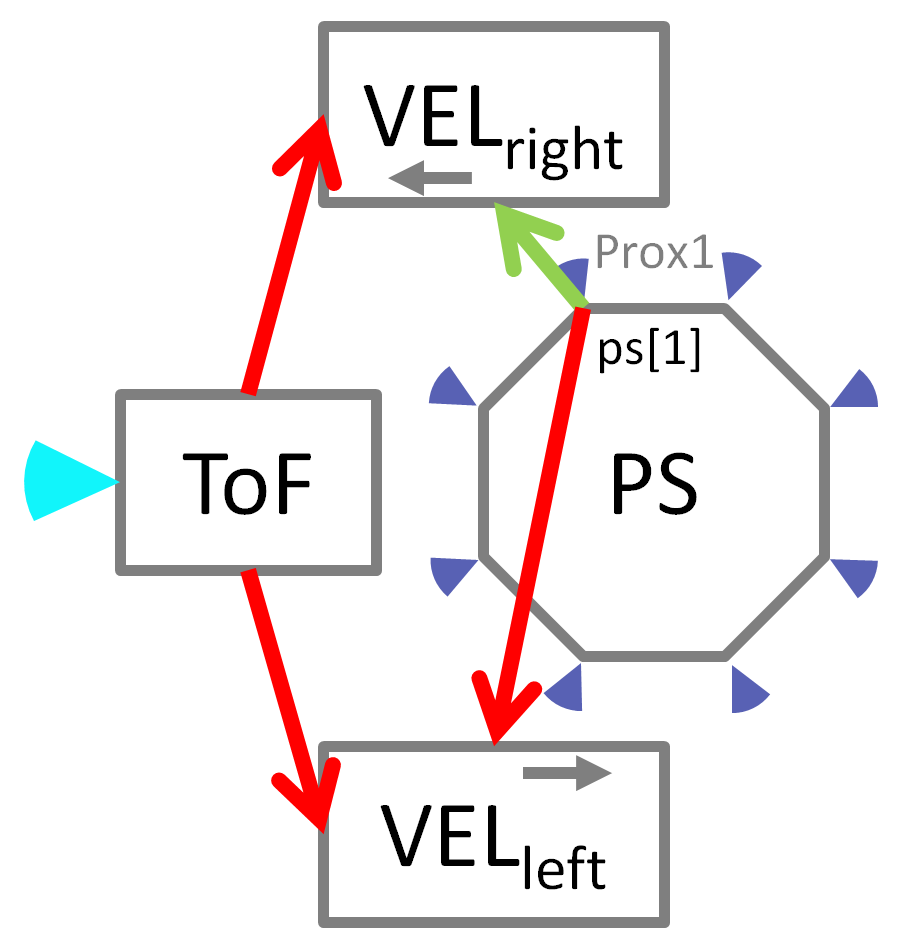}\label{subfig:schematic}} 
\caption{
Connecting sensory and motor neurons. 
(a) A minimalistic neuromorphic brain that controls the E-Puck robot.
Its layout was inspired by the topography of the motor and sensory cortex.
The blue markers indicate similar positions of the sensory neurons to the proximity sensors shown in Fig. \ref{subfig:EPuckSensors}.
The motor neurons are located close to the sensory neurons and at the corresponding side.
(b) Schematic of sensory neurons connecting to the motor neurons. If an obstacle comes in front of E-Puck, the ToF sensory neurons excite the backward motor neurons in VELleft and VELright and E-Puck slows down, respectively stops. Otherwise, the population-rate coding of the two sensory neurons in the neuron group ToF (ctx.tof) are configured in such way that they drive E-Puck forward. Fig. \ref{fig:io} shows E-Puck detecting a wall by its front-right proximity sensor Prox1 and the firing of the corresponding sensory neuron ps[1] in the neuron group PS (ctx.ps). The exemplary proximity sensor neuron ps[1] excites the proximal forward motor neuron of VELright and the distal backward motor neuron of VELleft making E-Puck turning left and therefore away from the wall. All of the eight proximity sensory neurons are connected in such a way. This enabled E-Puck to explore a T-shape maze without getting stuck in corners or passages.         
}
\label{fig:connections}
\end{figure*}

CARLsim's GUI provides methods for model creation, execution, visualization, and analysis. It contains plug-ins with a linkage to the CARLsim APIs that can be selected from a drop-down menu list. 
This allows the user to generate an SNN network with neuron groups of a specific type and the connectivity between groups. 
Figs. \ref{fig:neurons} and \ref{fig:connections} show how the CARLsim GUI can be used to generate a sensory-motor SNN for controlling the E-Puck robot (see Fig. \ref{fig:EPuckWall}). 

The network is visualized in 3D allowing browsing operations such as rotation, translation, zoom in/out, filtering, and selecting single neurons to follow neural pathways. 
The models are stored in a relational database (MySQL) which allows fast loading and even direct SQL queries.   
The user can start, pause, single-step, continue, and stop the execution of the SNN model in CARLsim to get direct feedback of its neural activity. 
CARLsim tools, such as its OAT and spike generators, are integrated as Qt-Widgets (plug-ins) named Monitors, and Injectors (Fig. \ref{subfig:gui_oat}). 
Fig. \ref{subfig:oat_plot} shows the spikes from the sensors and the resulting motor neurons that drive the actuators in the Monitor.  
The I/O interface is configurable over XML files and its devices listed with additional status information.


Using the GUI, without any text line programming, the SNN and controls were created such that the robot could explore a maze without getting stuck or colliding into walls (see Fig. \ref{fig:io}).
Since its sensory neurons were directly connected to the motor neurons, the robot can be thought of as a `neuromorphic' Braitenberg vehicle \cite{Braitenberg1984}. 
The SNN could readily be extended with interneurons, a neuromodulatory system, plasticity, and other sensors such as the camera or microphones (see \cite{Niedermeier2022} for examples).  

The visualization of the neural activity is useful for debugging and presentation purposes. Fig. \ref{fig:io} shows the firing of a neuron in the SNN. 
CARLsim GUI has a Time Panel, see Fig. \ref{fig:io} top left,   
that can increase the execution time to allow a more fine granular observation in slow motion. 
It also can accelerate 
the model execution the for faster training. 
Consequently, the time panel displays the different time lines such as model time versus real time. 



As with all components of the CARLsim system, the CARLsim GUI is fully open source.   
In order to deal with the complexity and to be extensible by its users,
the CARLsim GUI is designed as a highly modular Plug-In framework (see also Section \ref{sec:methods} Methods). 
It provides well defined interfaces 
for almost all components including the CARLsim simulator itself. 
The provided plug-ins also serve as templates and reference implementations. 
%
%
The extensibility of the plug-in framework provides a technical foundation expansion. Since the integration is by APIs, plug-ins can also be provided as binaries or used as such.



\subsection{CARLsim I/O - Input and Output Interfaces}  

CARLsim I/O leverages YARP (see also Section \ref{sec:methods} Methods)
to encapsulate robot controllers and other active nodes (e.g., datasets) as devices. Their sensors and actuators are published as ports that can be arbitrarily connected and are managed by a central name server. Communication protocols such as TCP/UDP or video compression are transparent to the user and can be easily configured. CARLsim I/O provides encoders to convert sensory data into spike streams and spike stream decoders into actuator commands.

Fig. \ref{fig:io} shows the I/O interface connecting the sensors and actuators of an E-Puck robot \cite{epuck2009} to an SNN that was built using CARLsim GUI (see Subsection \ref{subsec:results_gui}). The I/O interface was first configured as a digital twin of the E-Puck in Webots \cite{Webots2019,Webots2023}. After the test in the simulation environment, the YARP device plug-in was replaced by the physical E-Puck and the same SNN model operated the physical robot in a field test.  

\begin{figure*}[htb]
\centering
\subfloat[E-Puck starts exploring.]{\includegraphics[width=0.3\textwidth]{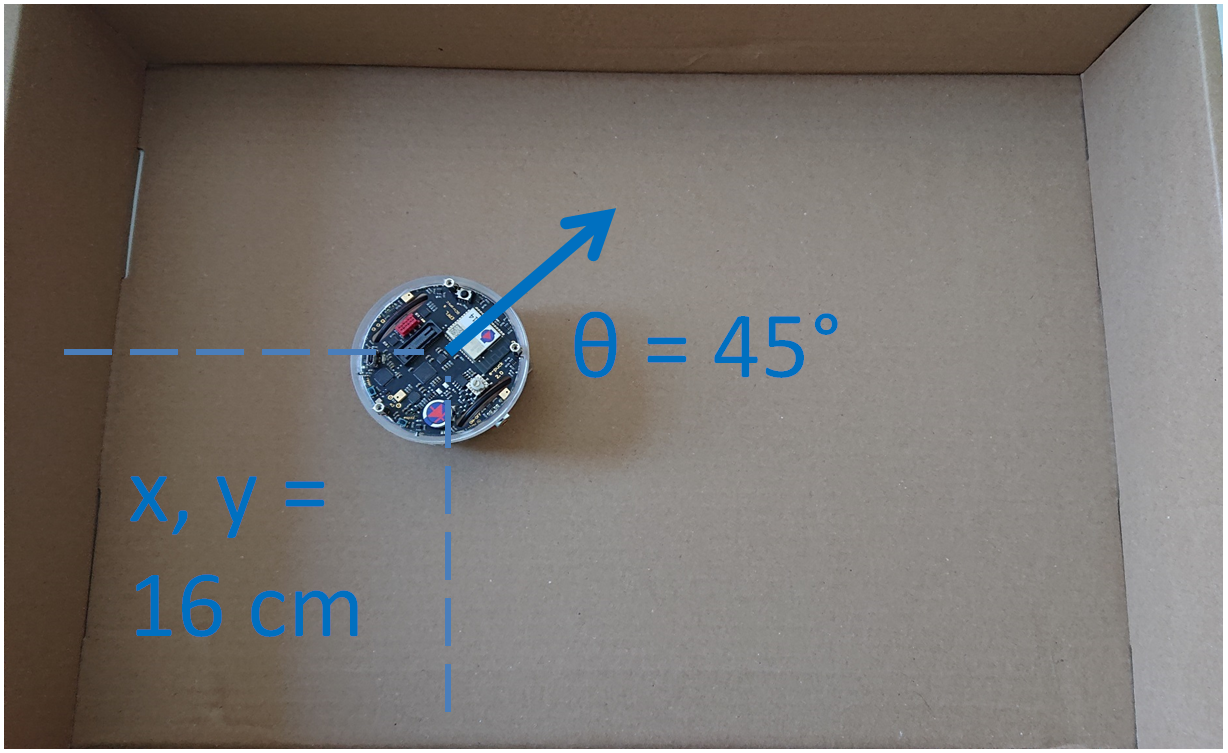}\label{subfig:experiment}} 
\hfil
\subfloat[Trajectory of collision avoidance.]{\includegraphics[width=0.35\textwidth]{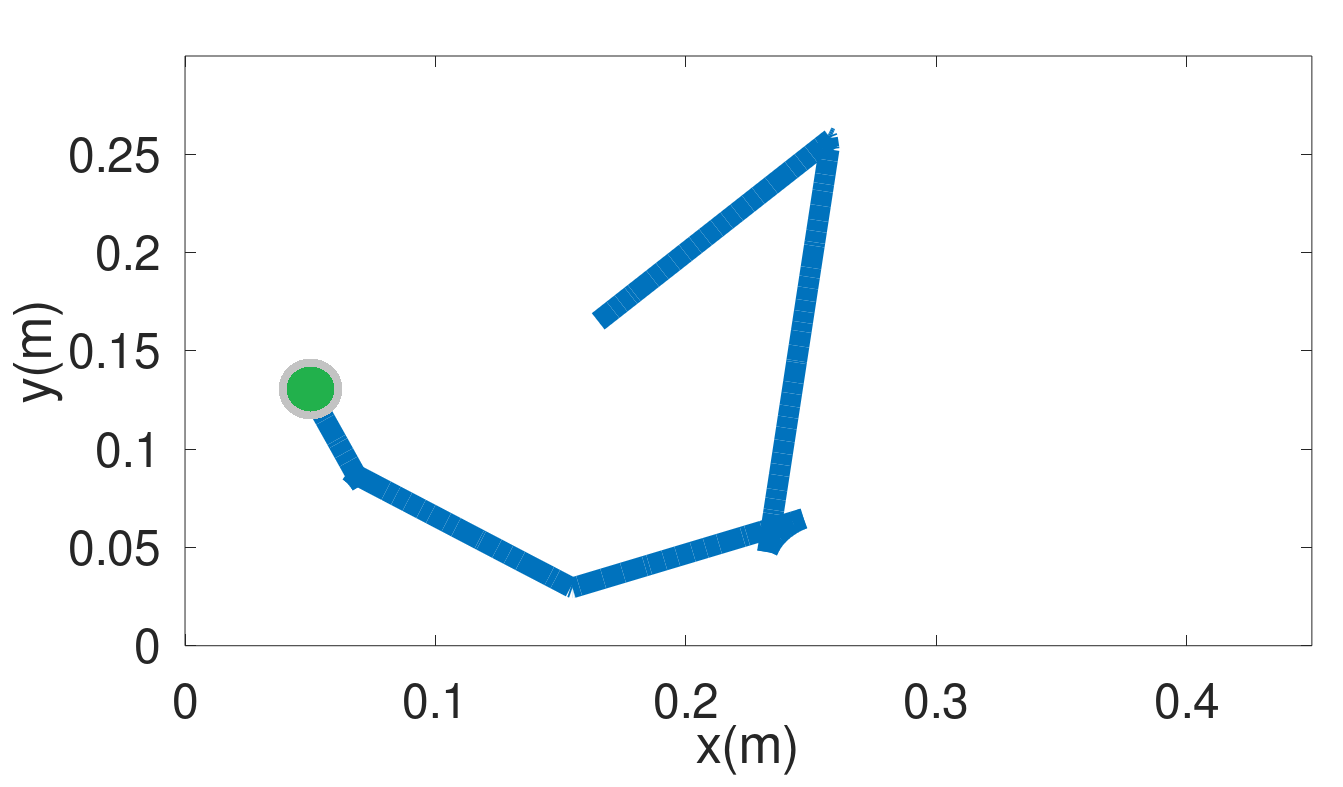}\label{subfig:trajectory}}  
\vfil
\subfloat[Plug-in for the CARLsim OAT]{\includegraphics[width=0.80\textwidth]{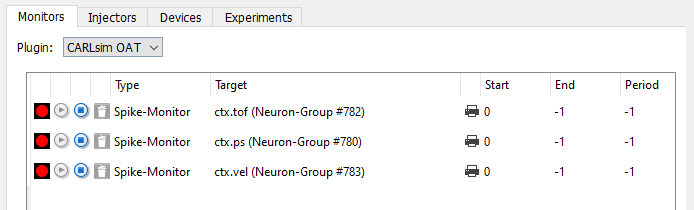}\label{subfig:gui_oat}}
\vfil
\subfloat[Neural activity of sensory and motor neurons]{\includegraphics[width=0.90\textwidth,height=2.0in]{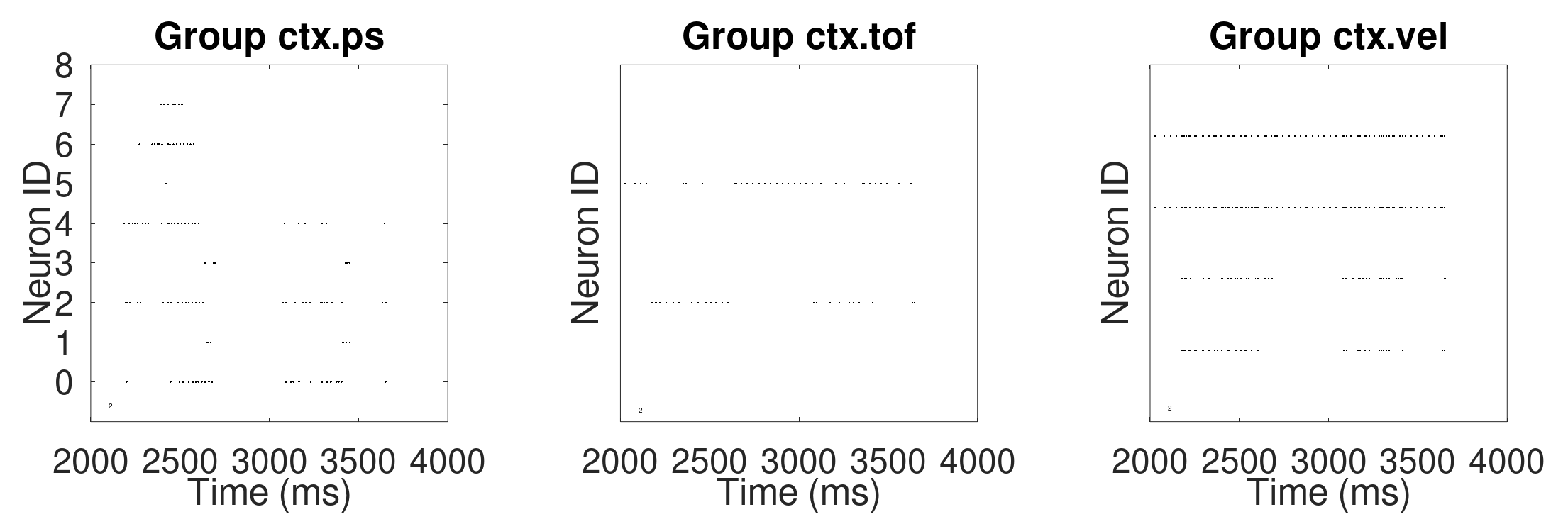}\label{subfig:oat_plot}}  
\caption{
Experiment showing E-Puck exploring an area restricted by walls and avoiding collisions. 
(a) Start position of E-Puck.
(b) Trajectory showing E-Puck exploring the field and changing course when coming near to a wall. 
Analysis of neural activity and interactive control of the monitoring. 
(c) The CARLsim++ plug-in for CARLsim OAT, which allows to start and stop an OAT monitor 
and to interactively access its results.
(d) Spike trains of neuron groups for sensors and actuators. The plots were generated by OAT and show the neuron group names as they were configured in CARLsim GUI. The namespace \emph{ctx} indicates that the groups belong to the motor and sensory cortex.
The time for the plot was arbitrary chosen to indicate high neural activity. At about 2.5s, several proximity sensor neurons were spiking, indicating that E-puck was near or in a corner. Consequently, ToF was detecting obstacles and the motor neurons encoded no forward movement. At about 2.8s, the silence of the ctx.ps neurons and the lower ctx.tof and ctx.vel neurons indicate that E-Puck was moving straight ahead. 
}
\label{fig:trajectory}
\end{figure*}

Fig. \ref{fig:trajectory} shows the trajectory showing E-Puck exploring an
area restricted by walls and avoiding collisions.
E-Puck's wheel sensor provides the total count of motor steps with a resolution of 1000 steps per revolution. 
We calculated the trajectory with the equation \ref{eq:Laymond} 
assuming nonholonomic constaint (no wheel slip) following \cite{Ibrahim2016} and \cite{Laumond1998}
\begin{eqnarray} \label{eq:Laymond}
\begin{pmatrix} \dot{x} \\ \dot{y} \\ \dot{\theta} \end{pmatrix} &{}={}& \begin{pmatrix} \cos\theta & 0 \\ \sin\theta & 0 \\ 0 & 1 \end{pmatrix} \begin{pmatrix} v \\ w  \end{pmatrix}
\end{eqnarray}  
with $v = \frac{1}{2}(v_l + v_r)$ and $w = \frac{1}{l}(v_l - v_r)$.
E-Puck's wheels have a diameter of 42mm ($r = 0.0215\text{m}$) and a distance of 55mm ($l = 0.055\text{m}$). 
The wheel velocities at a sensor event are $v_{l|r} = \frac{ steps_{l|r} }{500} \pi r / dt $.   
Equation \ref{eq:sum} was used to link the sensor events (in average every $dt = 128\text{ms}$),
from the start at $x_0 = y_0 = 0.16\text{m}$ and with a heading of $\theta_0 = \frac{1}{4}\pi$ (45°).

\begin{eqnarray} \label{eq:sum}
\begin{pmatrix} x_t \\ y_t \\ \theta_t \end{pmatrix} &{}={}& \begin{pmatrix} x_{t-1} \\ y_{t-1} \\ \theta_{t-1} \end{pmatrix} + \begin{pmatrix} \dot{x} \\ \dot{y} \\ \dot{\theta} \end{pmatrix} dt						
\end{eqnarray}  

An alternative method to estimate the trajectory is SwissTrack \cite{Lochmatter2008}
that utilizes a video camera to determine the robot position. 
We also provide short videos showing E-Puck2 in the experiment \cite{EPuckVideoPhys}
and simulated environments \cite{EPuckVideoWebots}. 

\subsection{CARLsim++ applications}

The YARP module manager provides a GUI to configure, start, and stop the CARLsim++ application visually.  
The configuration of the plug-ins and YARP devices are stored as versioned XML files. 
They are modular, so that for instance the same configuration file is used for both the physical E-Puck robot and its digital twin.
%
%
%
%
%
%
To run the digital twin of E-Puck, its simulation environment webots.exe starts automatically
on the YARP device ports prefixed by /wb, for example /wb/e-puck2/sensors instead of /e-puck2/sensors.
This means the same neuromorphic model 
including its monitoring configuration can be run with the physical device. 
This is critical for making a synergistic application evolution cycle possible. 
%
%
%
%
%
%
%
%

This modular configurability enables the neccessary unit and pre-integration testing. 
For instance, we used the EPuck monitor shown in Fig. \ref{subfig:EPuckMonitor} for validating the sensory data received from E-Puck. 
For our YARP device of the physical E-Puck, we developed a custom WinSock connector that establishes a TCP/IP link over WiFi on a configured IP address. 
With the YARP devices utility \emph{yarpdev}, 
we can then easily validate the data from EPuck's Time-of-Flight (ToF) sensor, 
that can measure the distance to frontal obstacles up to 2m. 

\subsection{Building applications}

The CARLsim++ Superbuild is a YCM reference configuration that allows a clean build from scratch. 
This means pure open source across all libraries, even down to MySQL driver and the latest Qt 6.x. 
In particular, the GIT repositories are cloned from GitHub, Bitbucket, and GitLab on demand. 
%
YCM automatically generates a dependency graph. 
We also provide templates and preconfigured YARP devices for custom applications.

CARLsim++ runs on Linux and Microsoft Windows. Microsoft, NVIDIA, and Ubuntu teamed together and 
position Windows Subsystem for Linux 2 (WSL2) as 
preferred compile and build development for cross-platform CUDA applications. 
With Windows 11, WSL2 fully supports GUI and utilizing the host CUDA drivers. 
As a consequence all YARP GUI applications run as nicely on WSL2 as the run on the native Windows host. 
By end of 2023, Microsoft has integrated CMake in Visual Studio 2022, 
one of the most sophisticated integrated development environment (IDE) for years.
Linux applications can now be fully developed and debugged in Visual Studio. 
An embedded emph{rsync} process synchronizes the source files between the Windows host
and the Linux instance (WSL2 or remote via SSH). 
In particular, VisualStudio interfaces with the GNU debugger running the Linux process. 
Also YARP can run distributions on both operating systems. 
For example the Linux instance can run the YARP infrastructure like name server and logger
and Windows can be used for YARP devices requiring proprietary drivers.


\subsection{Deployment and Cloud Native}

As outlined above, it is crucial to be able to display a neuromophic application 
in the simulation environment such as Webots and as physical field application
with minimal configuration and model changes. 
The binaries are generated cross-platform. 
For the deployment as Cloud Native applications or at the Edge, 
Docker is the defacto standard. 
We provide ready to use Docker images on our GitHub repository.
The corresponding Docker files are provided as well 
and allow to customize the Docker images for individual purposes. 

\subsection{DevOps - Running the application}  

\emph{DevOps} is today's industry standard. 
It denotes the continuous improvement by further \emph{Dev}elopment 
and the \emph{Op}eration\emph{s} of an application as a product. 
Operations require such tasks as monitoring, incident management,
and deployment. 
User feedback leads to improvements through Development. 
Configuration has to be under strict version control.

\subsubsection{Logging}

Logging is important for development and must be activated at runtime
and provide sufficient information of all components and their integration. 
The example below indicates how the distinct information are provided by CARLsim++.
The details of spike stream decoding a beyond of this work. 
The logfiles are collected in the /var/log directory 
as configured in the device XML, 
and have the sensor or actuator name as prefix, here \emph{vel\_}.

%
%

\subsubsection{Monitoring}

The CARLsim GUI presents neural activity upon the structure of the SNN in an intuitive way
that especially supports the user in the early modelling phase.   
For the quantitative analysis of the running neuromorphic application, 
neuroscientists utilize the CARLsim Offline Analysis Tool (OAT)   
which can generated Matlab figures. 
CARLsim++ integrates OAT as plug-in that interactive way. 
Fig. \ref{subfig:oat_plot} shows the neural activity of sensory and motor neurons
in Matlab figures generated by OAT.

\subsubsection{Robot Test Framework}  
The Robot Test Framework (RTF), a supplemental of YARP 
allows the quality control of more complex neuromorphic applications.
Its test cases can be interactively developed in Jupyter Lab. 
The tests can be automated using Python and, in particular, PyCARL library. 


%
%

\section{Conclusion} 

CARLsim++ provides a new approach and a framework for efficiently developing neuromorphic applications. Its core component is an efficient SNN simulator that is matured and rich on advanced features. Through the component CARLsim I/O, a neuromorphic robot can interact with its environment.  With the CARLsim GUI, users can easily build neuromorphic applications without programming (C++, Python). 
Users can configure and operate CARLsim monitors without having to write Matlab/C++ code as in previous CARLsim releases.

In future projects we will integrate more features of E-Puck, other robots, and other devices.  With the E-Puck, we plan to utilize the video feed from the camera for object recognition and the audio stream from the microphones will used for location of the sound source as in \cite{schoepe2019}. We plan a hybrid approach to large language models available as cloud service for processing of voice and text to speech.  We believe that CARLsim++ will be a useful framework for researchers and commercial developers who want to create neuromorphic applications.

\section{Acknowledgment}
This work was supported by the Air Force Office of Scientific Research (Contract No. FA9550-19-1-0306) and a UCI Beall Applied Innovation Proof-Of-Product award. We thank the regional NSF I-Corps program for their valuable insights. We also want to express our thanks to the active CARLsim community for their contributions.


\IEEEtriggeratref{1}
\IEEEtriggercmd{\enlargethispage{-0.1305in}}

\bibliographystyle{IEEEtran}
\bibliography{references}

\end{document}